\documentclass[letterpaper]{article}
\usepackage{proceed2e}
\usepackage[margin=1in]{geometry}
\usepackage[round]{natbib}
\usepackage[colorlinks=true,citecolor=blue,linkcolor=blue,urlcolor=blue]{hyperref}
\usepackage{graphicx}
\usepackage{svg}
\usepackage{epstopdf}
\usepackage{subfig}
\usepackage[]{bm}
\usepackage[]{amsmath}
\usepackage{amssymb}
\usepackage{mathrsfs}
\usepackage{multirow}% http://ctan.org/pkg/multirow
\usepackage{booktabs}% http://ctan.org/pkg/booktabs
\usepackage[colorinlistoftodos]{todonotes}
\usepackage{lipsum} % for dummy text only

\usepackage{times}

\title{Capturing Structure Implicitly from Time-Series having Limited Data}

\author{} % LEAVE BLANK FOR ORIGINAL SUBMISSION.
\author{ {\bf Daniel Emaasit\thanks{Corresponding author.}} \\
Haystax Technology \\
McLean, VA    \\
demaasit@haystax.com\\
\And
{\bf Matthew Johnson\thanks{Work done prior to joining Ibotta, Inc. Current contact is mjohnson@ibotta.com.}}   \\
Haystax Technology \\
McLean, VA    \\
mjohnson@haystax.com\\
}

\begin{document}

\maketitle

\begin{abstract}
Scientific fields such as insider-threat detection and highway-safety planning often lack sufficient amounts of time-series data to estimate statistical models for the purpose of scientific discovery. Moreover, the available limited data are quite noisy. This presents a major challenge when estimating time-series models that are robust to overfitting and have well-calibrated uncertainty estimates. Most of the current literature in these fields involve visualizing the time-series for noticeable structure and hard coding them into pre-specified parametric functions. This approach is associated with two limitations. First, given that such trends may not be easily noticeable in small data, it is difficult to explicitly incorporate expressive structure into the models during formulation.  Second, it is difficult to know \textit{a priori} the most appropriate functional form to use. To address these limitations, a nonparametric Bayesian approach was proposed to implicitly capture hidden structure from time series having limited data. The proposed model, a Gaussian process with a spectral mixture kernel, precludes the need to pre-specify a functional form and hard code trends, is robust to overfitting and has well-calibrated uncertainty estimates.
\end{abstract}

\section{INTRODUCTION}

Modeling expressive structure in time-series data is a vital requirement in several scientific fields including cyber security, insider-threat detection, and highway-safety planning for the purpose of scientific discovery. Chief risk officers (CRO) in large corporations and government agencies are tasked with forecasting potential risks to company assets by malicious employees or contractors. Departments of Transportation (DOTs) are mandated by the Federal Highway Administration (FHWA) to forecast highway crashes as part of their strategic highway safety plans (SHSPs) in order to obtain funding \citep{smith2016hsip}. However, these fields often lack sufficient amounts of training data to accurately capture the signal of the phenomenon under study and consequently discover hidden patterns. For instance \cite{greitzer2013methods} state that "ground truth" data on actual insider behavior is typically either not available or is limited. In some cases, one might acquire real data, but for privacy reasons, there is no attribution of any individuals relating to abuses or offenses—i.e., there is no ground truth. The data may contain insider threats, but these are not identified or knowable to the researcher \citep{greitzer2013methods,gheyas2016detection}. In highway-safety planning, \cite{veeramisti2016business} mentions that Departments of Transportation (DOTs) only recently started collecting monthly highway-crash data because of the high cost and extensive process of collecting the required data.

\subsection{INSIDER THREAT}\label{insider-threat}

Insider threats are malicious acts carried out by current or former employees or trusted partners of an organization who abuse their authorized access to an organization's networks, systems, and/or data \citep{glasser2013bridging, lindauer2014generating}. Insider threats include but not limited to, theft of intellectual property or national security information, fraud, and sabotage. Many government, academic, and industry groups seek to discover and develop solutions to detect and protect against these insider threats. However, the difficulty of obtaining suitable data for research, development, and testing remains a significant hinderance \citep{glasser2013bridging}. This is attributed to two major factors. First, confidentiality and privacy concerns create barriers to the collection and use of such data for research purposes. To collect real data, some organization must directly monitor and record the behavior and actions of its own employees. Second, insider threats are black swan events that are very rare and only collected after the fact \citep{gheyas2016detection}. Nonetheless, some studies have proposed methods for time-series analysis. For example, \cite{stoffel2013finding} proposed a  Fourier and wavelet model, to estimate a time-series model for internet-traffic data from a computer network of workstations and servers.

\subsection{HIGHWAY SAFETY}\label{highway-crash-safety}

The Federal Highway Administration (FHWA) requires state Departments of Transportation (DOTs) to develop Highway Safety Plans (SHSPs) to obtain funding as part of the two legislative acts \citep{smith2016hsip}. State Departments of Transportation (DOTs) in the United States are tasked with making reasonable predictions of highway crashes on roadways based on historical data. This is so as to set realistic targets for performance-based safety programs to reduce fatalities and serious injuries. For a traffic safety policy, a model-based approach generally is avoided for forecasting and setting targets \citep{veeramisti2016business}. This is because the required data collection is extensive and expensive; in addition, establishing a relationship between the performance measures and influencing factors is difficult \citep{kweon2012appropriate}. For example, in order to use a model-based approach, the Average Annual Daily Traffic (AADT) is one of the exposure variables that is required for all public roads. Most DOTs only collect the corresponding data for a sample of their facilities \citep{veeramisti2016business}. Nonetheless, some studies have proposed model-based methods for time-series analysis. For example, \cite{veeramisti2016business} proposed a stochastic time series model, a seasonal auto-regressive integrated moving average (sARIMA) model, to forecast performance measures for performance-based safety programs for the Nevada DOT. Other studies have proposed different variations of ARIMA for crash forecasting \citep{yannis2011modeling,sukhai2011temporal}. These targets can be used to determine future statewide safety improvement programs and policies. From the perspective of state agencies, predicting the number of fatalities and serious injuries is significantly important to meet the requirements of MAP-21 \citep{smith2016hsip}.

\subsection{LIMITATIONS AND RESEARCH QUESTION}\label{limitations-and-research-questions}

Current methods for time-series modeling for insider-threat detection and highway-safety planning (described in sections \ref{insider-threat} and \ref{highway-crash-safety}) are associated with two major limitations.  First, the methods involve visualizing the time series for noticeable structure and patterns such as periodicity, smoothness, growing/decreasing trends and then hard coding these patterns into the statistical models during formulation. This approach is suitable for large datasets where more data typically provides more information to learn expressive structure \citep{wilson2014covariance}. That is, an analyst is able to look at the data, recognize trends (i.e. periodicity, seasonal variation with possible decay away from periodicity, long-term rising/decreasing trends, medium-term irregularities, short seasonal variations, and noise), and then hard code those patterns into statistical algorithms such as ARIMA to represent each of these patterns. The limitation with this approach is that all the pattern discovery is done by the analyst and the resulting algorithm is used simply as a smoothing tool. Given limited amounts of data, it is difficult to explicitly incorporate expressive prior information (such as periodicity, smoothness, growing tends) into the statistical learning algorithms. 

Second, most of the current literature focus on parametric models that impose strong restrictive assumptions by pre-specifying the functional form and number of parameters \citep{veeramisti2016business,gheyas2016detection}. Pre-specifying a functional form for a time-series model could lead to either overly complex model specifications or simplistic models. It is difficult to know \textit{a priori} the most appropriate function to use for modeling sophisticated insider-threat behavior and highway-crash occurrence that involve complex hidden patterns and many other influencing factors. Due to a finite number of parameters, parametric models are not necessarily the best to capture the entire structure of complex and sophisticated processes \citep{ghahramani2015probabilistic}, such as insider-threat behavior and highway-crash occurrence. 

Given limited and noisy time-series data for insider-threat detection and highway-safety planning, is it possible to perform: (1) pattern discovery without hard-cording trends into statistical models during formulation, and (2) model estimation that precludes pre-specifying a functional form? To answer these two research questions and address the above described limitations, this paper proposed a nonparametric Bayesian framework which is applied as an example to estimate two time series models: (1) the size of email attachments sent outside an organization by insiders and (2) the annual number of highway crashes in Nevada. The proposed framework combines the flexibility of nonparametric Bayesian approaches \citep{williams2006gaussian,hjort2010bayesian,ghahramani2015probabilistic} and kernels with more expressive properties \citep{wilson2013gaussian, duvenaud2014automatic} than standard kernels proposed by \cite{williams2006gaussian}. In particular, a flexible nonparametric model -- i.e., the Gaussian process (GP) model --  is proposed as a prior distribution for the unknown regression functions, thereby precluding the need to pre-specify a functional form. Flexibility is achieved by making weaker assumptions about model structure, thereby capturing the underlying behavior from data by exploring an infinite dimensional space of possible functions. Moreover, an expressive covariance -- i.e., the spectral mixture (SM) kernel \citep{wilson2014covariance} -- is proposed as the covariance for the GP model thereby precluding the need to hard cord trends and consequently allows to automatically discover hidden structure from limited data. The SM kernel is particularly useful when standard kernels such as the radial basis function fail to capture functional structure \citep{schulz2017compositional} in limited noisy data commonly found in insider-threat detection and highway-crash analysis. Recent advances in computing and efficient sampling methods, such as the Hamiltonian Monte Carlo (HMC) \citep{hoffman2014no} enable fast inference in flexible models. Moreover, a Bayesian modeling approach was used for inference to quantify all uncertainty, using probability distributions. The proposed GP-SM model maintained the attractive interpretability properties of a standard GP model; that is, the posterior mean and credible intervals of the predicted series could be analyzed to visualize how the response variable varies over time. 

The rest of this paper is organized as follows. Section \ref{methodology} describes the proposed methodology, including model formulation and estimation using two efficient Bayesian inference methods. Section \ref{experiments} describes the experiments performed including a description of the data used, the sample formation processes, and empirical analyses. Finally, a conclusion that summarizes the important findings from this study are presented in Section \ref{conclusions-and-future-work}, including recommendations for future research.
 
\section{METHODOLOGY}\label{methodology}

\subsection{MODEL FORMULATION}\label{model-formulation}

To formulate the methodology, consider for each data point, $i$, that $y_i$ represents a response variable (such as the attachment size in emails sent by an insider outside the organization or number of highway crashes) and $x_i$ is a temporal covariate such as year, month, week, or day. Regression modeling can involve estimating a latent function \(f\), which maps input data, $x_i$, to output data \(y_i\) for \(i\) = 1, 2, \ldots{}, \(N\), where \(N\) is the total number of data points. Each of the input data $x_i$ is of a single dimension $D = 1$, and \(\textbf{X}\) is a \(N\) x \(D\) matrix with rows $x_i$. The observations are assumed to satisfy:
\begin{equation}\label{eqn:additivenoise}
y_i = f(x_i) + \varepsilon, \quad where \, \, \varepsilon \sim \mathcal{N}(0, \sigma_{\varepsilon}^2)
\end{equation}
The noise term, $\varepsilon$, is assumed to be normally distributed with a zero mean and variance, $\sigma_{\varepsilon}^2$. Latent function \(f\) represents hidden underlying trends that produced the observed time-series data.

Given that it is difficult to know $\textit{a priori}$ the most appropriate functional form to use for \(f\), a prior distribution, \(p(\bm{f})\), over an infinite number of possible functions of interest is formulated. A natural prior over an infinite space of functions is a Gaussian process prior \citep{williams2006gaussian}. Formally, a GP is defined as a collection of random variables, \(f(x_i)\), any finite subset of which, \(\textbf{f} = \{f(x_i)\}_{i=1}^N\), has a joint Gaussian distribution, \(p(\textbf{f} \mid \textbf{X}) = \mathcal{N}(\textbf{f} \mid \textbf{m}, \textbf{K}_{N,N})\). The parameter \(\textbf{m}\) denotes the mean function (or mean vector) which specifies the expected output of the function \(\textbf{f}\) given input $\textbf{X}$, and \(\textbf{K}_{N,N}\) denotes the covariance function (or covariance matrix or kernel) which specifies the covariance between outputs \(\textbf{y} = \{(y_i)\}_{i=1}^N\). A GP is fully parameterized by a mean function and covariance function, denoted as: 
\begin{equation}\label{eqn:gpsim}
\textbf{f} \sim \mathcal{GP}(\textbf{m}, \textbf{K}_{N,N}),
\end{equation}
where each element in the mean vector and covariance matrix is given by equation \ref{eqn:params-mean} and \ref{eqn:params-kernel}, respectively:
\begin{align}
m(x_i) &=\mathbb{E}[f(x_i)], \label{eqn:params-mean}\\
k(x_i, x_j) &=\mathbb{E}[(f(x_i)-m(x_i))(f(x_j)-m(x_j))^{\mathsf{T}}] \label{eqn:params-kernel}.
\end{align}
The kernel function, $\textbf{K}_{N,N}$, is used to encode prior assumptions of smoothness of functions such as periodicity, rising or decreasing long-term and short-term trends \citep{williams2006gaussian}. However, such properties may not be easily noticeable in time-series data for insider-threat detection and highway-crash forecasting. To address this limitation, a spectral mixture (SM) kernel \citep{wilson2014covariance,wilson2015human} was proposed for the covariance function. The SM kernel is able to implicitly capture structure by leveraging the idea that any standard kernel can be expressed as an integral using Bochner’s theorem, as illustrated below.

A standard stationary kernels is a functions of $\tau=x_i - x_j \in \mathbb{R}^p$, then
\begin{equation}\label{eqn:bochner-theorem}
k(\tau) = \int_{\mathbb{R}_P} \exp({2\pi is^\mathsf{T}\tau})\psi(d\mathbf{s}).
\end{equation}
If $\psi$ has a density $S(\mathbf{s})$, then $S$ is the spectral density of $k$; $S$ and $k$ are thus Fourier duals (\cite{williams2006gaussian}, see Appendix). This means that a spectral density over the kernel space fully defines the kernel and that furthermore every stationary kernel can be expressed as a spectral density. The spectral density modeled with a single Gaussian can be expressed as:
\begin{equation}\label{eqn:single-guassian}
\phi(s,\mu,\sigma^2) = \frac{1}{{\sigma \sqrt {2\pi } }}e^{{{ - \left( {s - \mu } \right)^2 } \mathord{\left/ {\vphantom {{ - \left( {s - \mu } \right)^2 } {2\sigma ^2 }}} \right. \kern-\nulldelimiterspace} {2\sigma ^2 }}}
\end{equation}
The density $S(\mathbf{s})$ is called the spectral density of $k$ and is given by:
\begin{equation}\label{eqn:single-guassian}
S(\mathbf{s})=\frac{1}{2}[\phi(s)-\phi(-s)]
\end{equation}
The resulting kernel is given by:
\begin{equation}\label{eqn:spectral-kernel}
k(\tau)=\exp(-2\pi^2\tau^2\sigma^2)\cos(2\pi\tau\mu)
\end{equation}
The spectral density can be approximated by a mixture of $Q$ Gaussians \citep{wilson2015human} as follows:
\begin{equation}\label{eqn:mixture-of-guassians}
k(\tau) = \sum_{q=1}^{Q}w_q\prod_{p=1}^{P} \exp(-2\pi^2\tau_{p}^2\upsilon_q^2)\cos(2\pi\tau_p\mu_q^{(p)}),
\end{equation}
where the $q$th component has mean vector $\bm{\mu}_q=(\mu_q^{(1)}, ...,\mu_q^{(p)})$ and a covariance matrix $\mathbf{M}_q$ = diag$(\upsilon_q^{(1)}, ..., \upsilon_q^{(p)})$. The inverse mean represents the component periods and the inverse standard deviation the length scales. The result is a flexible and expressive parametrization of the kernel, in which complex kernels are approximated by mixtures of simpler ones. The reader is refereed to \cite{wilson2015human} for a more detailed formulation of the SM kernel.

The likelihood of the response variable is a noisy sample from the latent function expressed as:
\begin{equation}\label{eqn:data-likelihood}
p(\textbf{y} \mid \textbf{f}, \textbf{X}) = \mathcal{N}(\textbf{f}, \sigma_{\varepsilon}^2).
\end{equation}
Given the GP prior in Equation \eqref{eqn:gpsim} and the data likelihood in Equation \eqref{eqn:data-likelihood}, the posterior distribution over the unknown function evaluations \(\textbf{f}\) at all data points \(x_i\), was estimated using Bayes theorem: 
\begin{equation}\label{eqn:bayesinfty}
\begin{aligned}
p(\textbf{f} \mid \textbf{y},\textbf{X}) &= \frac{p(\textbf{y} \mid \textbf{f}, \textbf{X}) \, p(\textbf{f})}{p(\textbf{y} \mid \textbf{X})}, \\
&= \frac{p(\textbf{y} \mid \textbf{f}, \textbf{X}) \, \mathcal{N}(\textbf{f} \mid \textbf{m}, \textbf{K}_{N,N})}{p(\textbf{y} \mid \textbf{X})},
\end{aligned}
\end{equation}
where:

\begin{center} 
\begin{minipage}{10cm} 
\begin{tabbing}
\phantom{$D_{n50}\ $}\= \kill
$p(\textbf{f}\mid \textbf{y},\textbf{X})$ = the posterior distribution of functions that\\ best explain the response variable, given the covariates \\
$p(\textbf{y} \mid \textbf{f}, \textbf{X})$ = the likelihood of response variable, given\\ the functions and covariates \\ 
$p(\textbf{f})$ = the prior over all possible functions of the\\ response variable \\
$p(\textbf{y} \mid \textbf{X})$ = the data (constant)
\end{tabbing}
\end{minipage} 
\end{center}
This posterior is a Gaussian process composed of a distribution of possible functions that best explain the time-series pattern. Given that the data are fixed, equation \eqref{eqn:bayesinfty} was re-formulated as the unnormalized posterior distribution
\begin{equation}\label{eqn:bayesinfty-unormalized}
p(\textbf{f} \mid \textbf{y},\textbf{X}) \propto p(\textbf{y} \mid \textbf{f}, \textbf{X}) \, \mathcal{N}(\textbf{f} \mid \textbf{m}, \textbf{K}_{N,N}).
\end{equation}
The posterior predictive distribution for a new input $x_{\star}$ conditional on observed data $\mathscr{D} = \{y_i, x_i\}_{i=1}^{N}$ is Gaussian with mean and variance given by:
\begin{align}
\mathbb{E}[f(x_{\star} \mid \mathscr{D})] &=k_{\star}^{\mathsf{T}}(\textbf{K}_{N,N}+\sigma^2\textbf{I})^{-1}\textbf{y}, \label{eqn:mean-predict}\\
\mathbb{V}[f(\textbf{x}_{\star} \mid \mathscr{D})] &=k(\textbf{x}_{\star}, \textbf{x}_{\star})-\textbf{k}_{\star}^{\mathsf{T}}(\textbf{K}_{N,N}+\sigma^2\textbf{I})^{-1}\textbf{k}_{\star} \label{eqn:kernel-predict}.
\end{align}
where $\textbf{k}_{\star}$ is the covariance between each observed input and the new input $\textbf{x}_{\star}$.

% \begin{table*}[t]
% \caption{Descriptive statistics of durations (in mins) by activity type}
% \label{tab:descriptive-stats-injury}
% \begin{center}
% \begin{tabular}[t]{lrrrrrr}
% \multicolumn{1}{c}{\bf ACTIVITY TYPE}  &\multicolumn{1}{c}{\bf MEAN}  &\multicolumn{1}{c}{\bf STD DEV}  &\multicolumn{1}{c}{\bf MIN}  &\multicolumn{1}{c}{\bf MAX}  &\multicolumn{1}{c}{\bf FREQUENCY}  &\multicolumn{1}{c}{\bf PERCENTAGE (\%)} \\
% \hline \\
% Personal activities & 309.5 & 269.0 & 1.0 & 1,370 & 12,296 & 51.4\\
% Meals & 342.1 & 262.1 & 1.0 & 1,249 & 4,837 & 20.2\\
% Shopping & 35.2 & 44.3 & 1.0 & 635 & 3,735 & 15.6\\
% Physically inactive recreation & 157.3 & 187.1 & 1.0 & 1,069 & 1,896 & 7.9\\
% Physically active recreation & 186.4 & 165.9 & 1.0 & 1,254 & 1,142 & 4.8\\
% \end{tabular}
% \end{center}
% \end{table*}

\subsection{MODEL ESTIMATION}\label{model-estimation}

Model estimation involved finding values for eighteen hyperparameters, including nine frequency parameters and nine lengthscale parameters. These hyperparameters describe the distribution over functions rather than the functions themselves. Bayesian inference, which was adopted to estimate these hyperparameters, involved expressing prior beliefs over all the hyperparameters and using probability theory to update those beliefs in light of the observed data. In this approach, in contrast to MLE, overfitting is of less concern because optimization or the minimization of an error is not used for estimation \citep{mchutchon2014nonlinear,ghahramani2015probabilistic}. Recent sampling methods considered to be efficient are used to explore the posterior distribution of the proposed model. Initial sampling was performed using an approximate method, known as automatic differentiation variational inference (ADVI) \citep{kucukelbir2015automatic}. Final sampling used an exact Markov chain Monte Carlo (MCMC) method, known as the Hamiltonian Monte Carlo (HMC); in particular, the No-U-Turn Sampler (NUTS) \citep{hoffman2014no} was used. Bayesian optimization \citep{brochu2010tutorial, emaasit2018simultaneous} was used to tune the hyperparameters.

Code and data for the experiments described in the next section are available on GitHub at \href{https://github.com/emaasit/long-range-extrapolation}{https://github.com/emaasit/long-range-extrapolation}.

\section{EXPERIMENTS}\label{experiments}

\subsection{INSIDER THREAT}
\subsubsection{Raw data and sample formation}

The insider-threat data used for empirical analysis in this study was provided by the computer emergency response team (CERT) division of the software engineering institute (SEI) at Carnegie Mellon University \citep{CERT}. The data, referred to as the "Insider Threat Tools dataset", is a synthetic test dataset generated by \cite{glasser2013bridging} that represents a fictitious company with employees and features such as their computers, files, removable media and websites they visited. Version "r6.2.tar.bz2" of the dataset was used. The particular insider threat focused on is the case where a known insider sent information as email attachments from their work email to their home email.

\begin{figure}[h]
\begin{center}
% \vspace{1in}
\includegraphics[width=\linewidth]{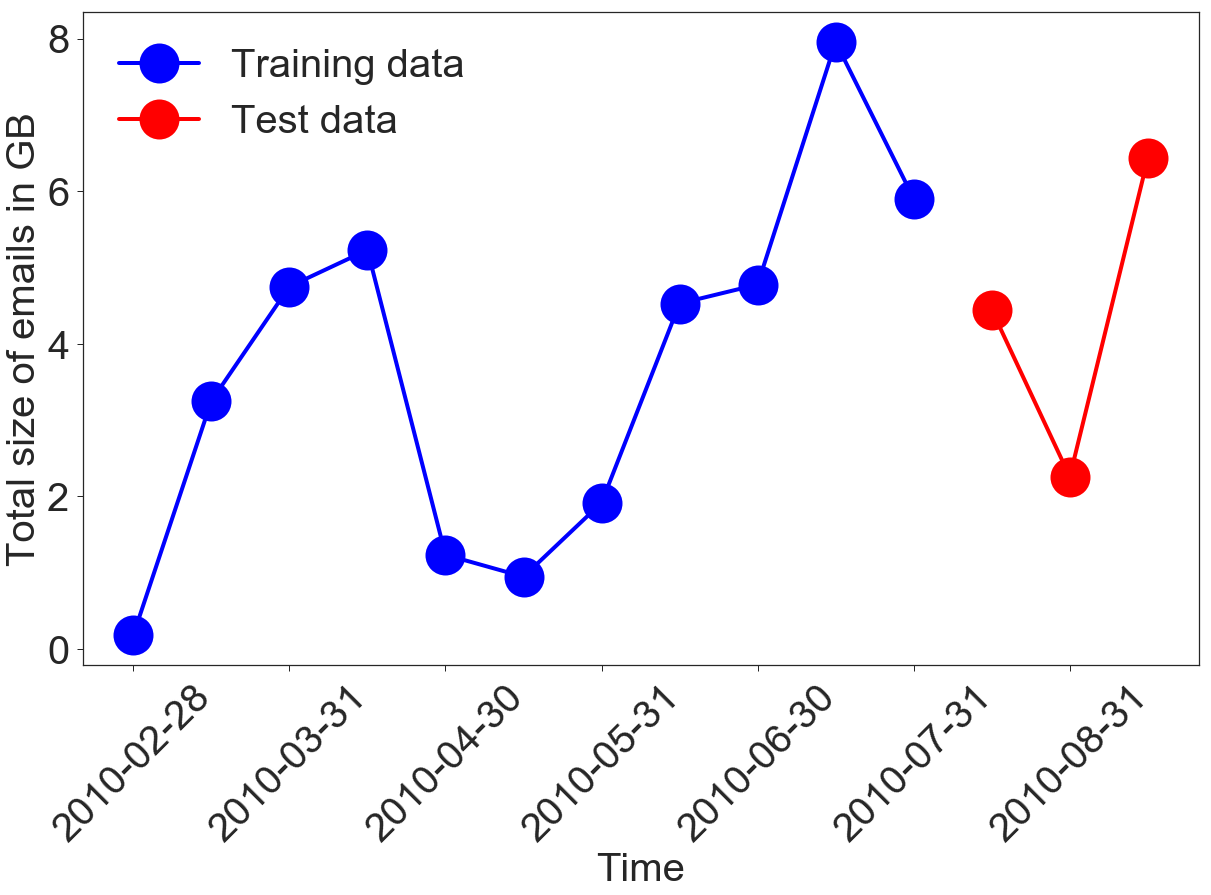}
\caption{Monthly-attachment-size of emails in GB sent by a known insider to a personal account.}
\label{fig:data-emails}
\end{center}
\end{figure}

To obtain the sample for the current analysis, the dataset was processed as follows. First, the data was filtered for a known insider with userID "CDE1846". Second, only emails sent by the insider outside the company organization were filtered. Third, the email-attachment size in Gigabytes was summarized into a monthly sum resulting into a sample size of 14 months ranging from 2010-01-31 to 2011-02-28. Table \ref{tab:descriptive-stats-emails} shows a comparison of descriptive statistics between emails sent to a personal account versus a non-personal account. The size of emails sent outside the company organization is shown to be smaller than amount sent within. Finally, emails sent to the insider's personal email account ("Ewing-Carlos@comcast.net") were filtered as shown in Figure \ref{fig:data-emails}. This was the final sample used for analysis. It is difficult to tell any noticeable trends in this small sample size of 14.
% \lipsum[1-1]

\begin{table}[h]
\caption{Descriptive statistics for monthly-attachment-size of emails in GB sent by a known insider outside the organization.}
\label{tab:descriptive-stats-emails}
\begin{center}
\begin{tabular}{l@{\qquad}cc}
  \toprule
  \multirow{2}{*}{\raisebox{-\heavyrulewidth}{\bf STATISTIC}} & \multicolumn{2}{c}{\bf EMAIL DESTINATION} \\
  \cmidrule{2-3}
  & {\bf Personal account} & {\bf Non-personal} \\
  \midrule
  Mean & 3.84 & 37.52  \\
  Std Dev & 2.28 & 13.38  \\
  Min & 0.18 & 10.56  \\
  Max & 7.96 & 57.78  \\
  Count & 14 & 14  \\
  \bottomrule
\end{tabular}
\end{center}
\end{table}

\begin{figure*}[t]
      \subfloat[GP-SM\label{subfig:model-emails}]{%
       \includegraphics[width=3.1in]{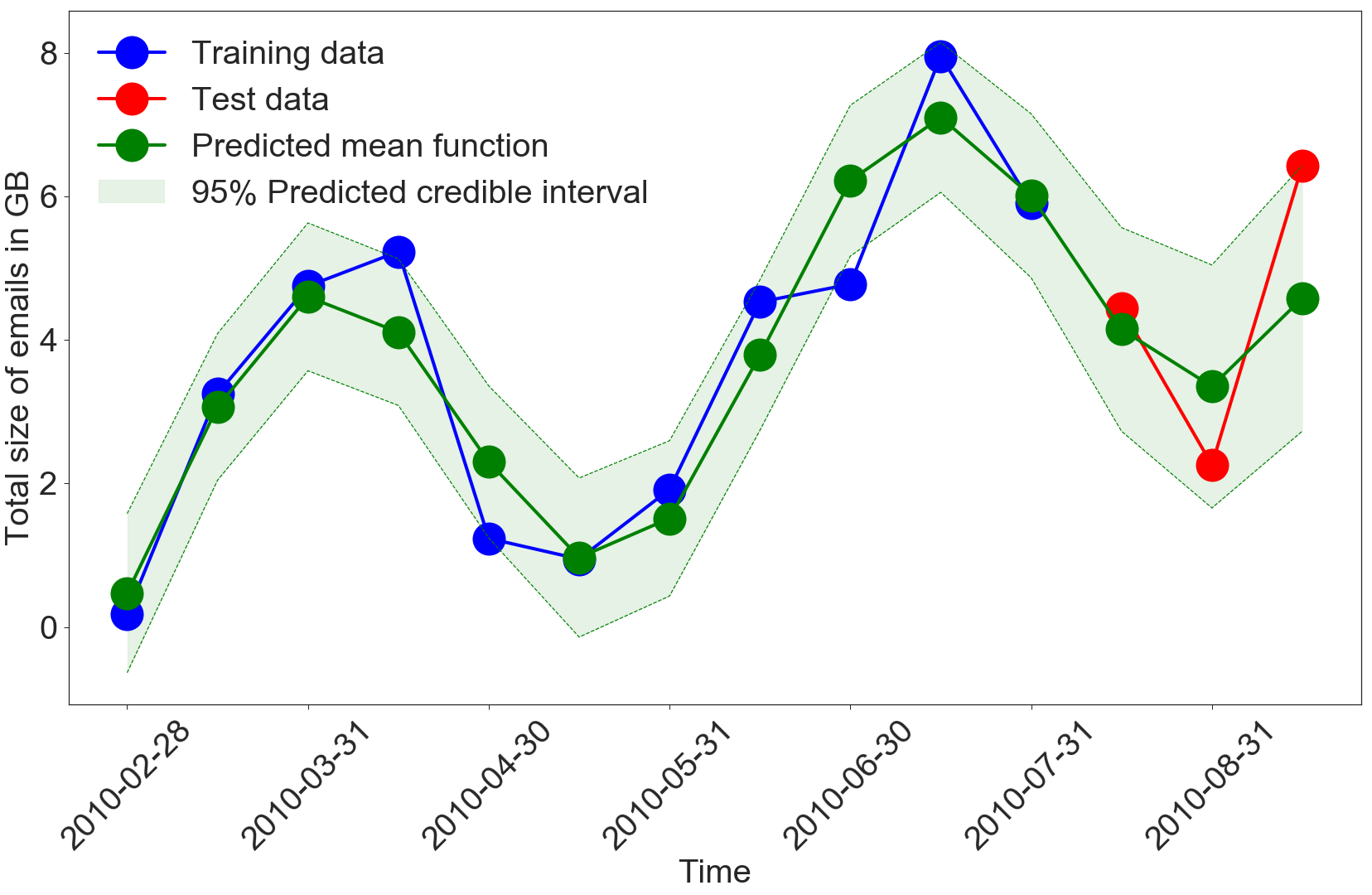}
     }
     \hfill
     \subfloat[GP-SM Optimized\label{subfig:model-opt-emails}]{%
       \includegraphics[width=3.1in]{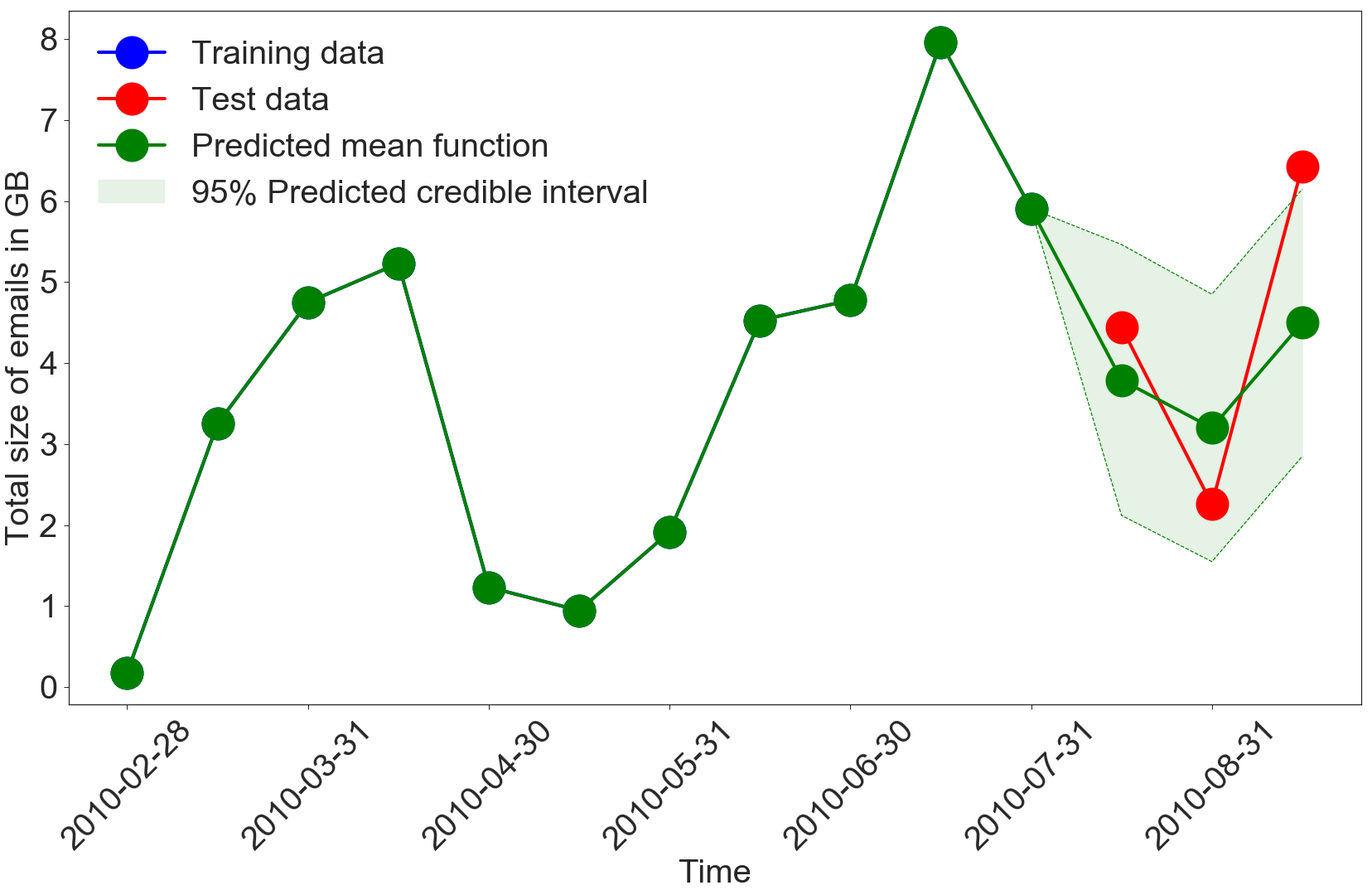}
     }
     \hfill
     \subfloat[ARIMA\label{subfig:model-emails-arima}]{%
       \includegraphics[width=3.1in]{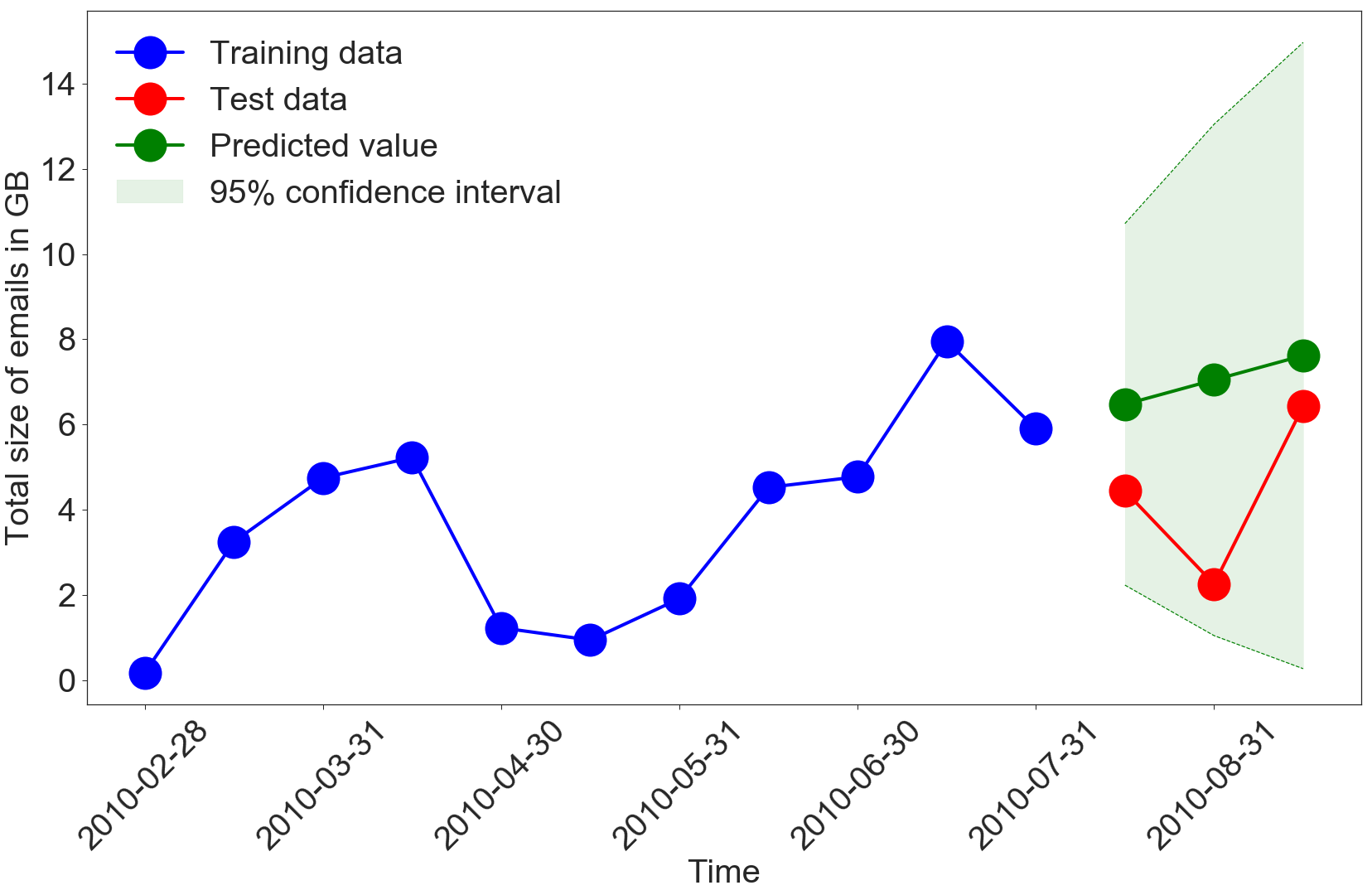}
     }  
     \hfill
     \subfloat[Number of Iterations of Bayesian Optimization\label{subfig:iterations-email}]{%
       \includegraphics[width=3.1in]{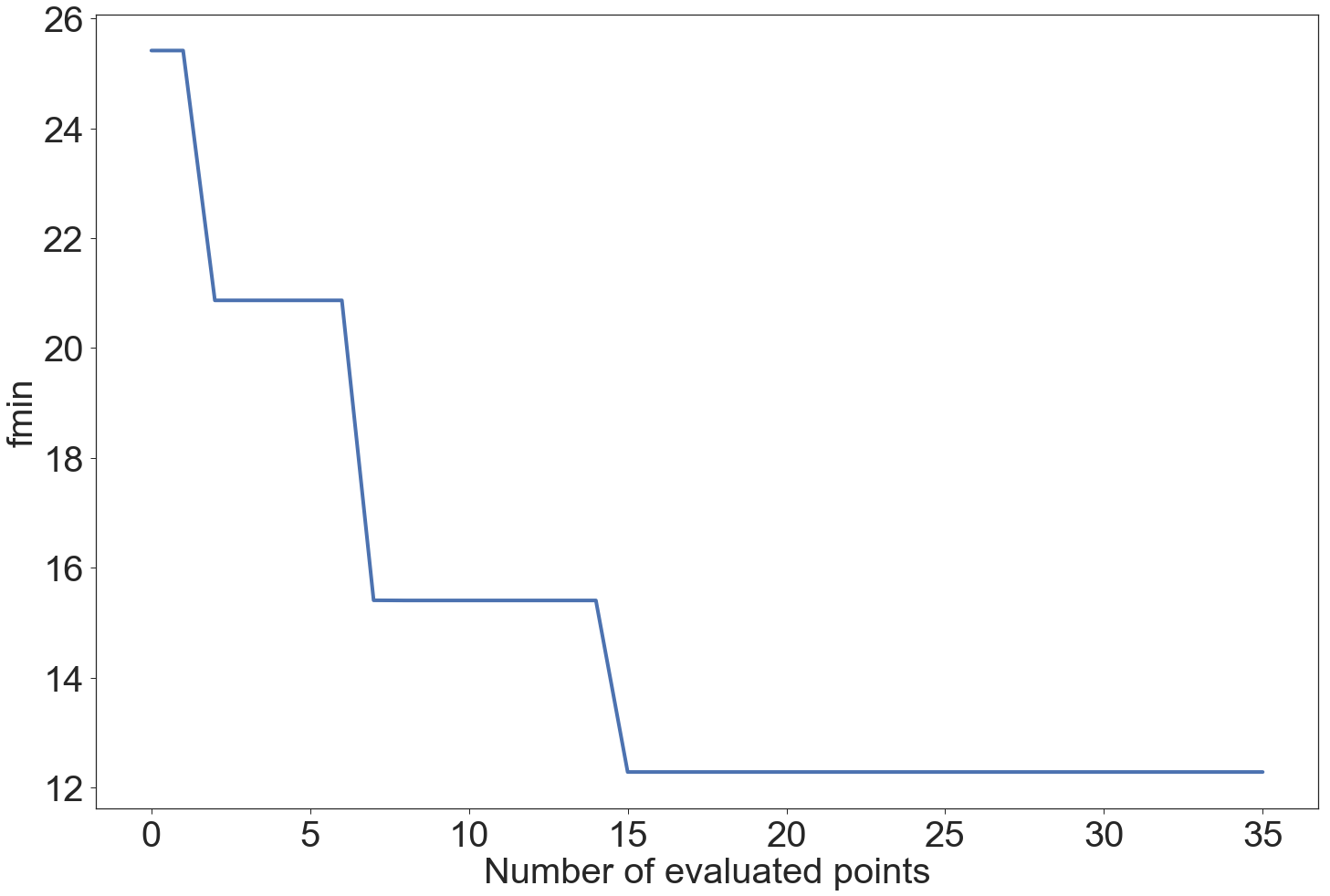}
     }   
     \caption{Comparison of Gaussian process models with spectral mixture kernels to ARIMA models for emails.}
     \label{fig:gp-posterior-emails}
\end{figure*}

\subsubsection{Empirical Analysis}
The first eleven data points shown in blue in Figure \ref{fig:data-emails} were used for training and the last three data points in red for testing. An appropriate number of terms in the sum, $Q$, was set to 10 as proposed by \cite{wilson2015human} resulting in 10 frequency parameters and 10 length-scale parameters. A maximum number of 30 iterations was chosen to tune the 20 hyperparameters for the optimized GP-SM. An ARIMA model was estimated using the methodology proposed by \cite{veeramisti2016business} for comparison.

Figure \ref{fig:gp-posterior-emails} shows the results of the three estimated models. Figure \ref{subfig:model-emails} shows that the Gaussian process model with a spectral mixture kernel is able to capture the structure implicitly both in regions of the training and testing data. The 95\% predicted credible interval (CI) contains the true size of email attachments for the duration of the measurements. Furthermore, the C.I are much narrower than the other two models. Figure \ref{subfig:model-opt-emails} shows that the Gaussian process model with the hyperparameters of the spectral mixture kernel tuned using Bayesian optimization is also able to capture structure implicitly. The predictive performance is especially better in the region of the training data where the predicted data points entirely overlap with the training data. Performance in the region of testing data is comparable to the standard model. This finding suggests that hyperparameter tuning for small data without easily noticeable structure does not produce any significant model improvement. Figure \ref{subfig:model-emails-arima} shows that ARIMA model is not able to capture the structure within the region of training data. This finding suggests that ARIMA models have poor performance for small data without noticeable structure. The 95\% confidence interval for ARIMA is much wider than both the GP models showing a high degree of uncertaininty about the ARIMA predictions.

Table \ref{tab:performance-measures-emails} shows the performance measures for the three estimated models including the root mean square error (RMSE) and mean absolute percentage error (MAPE). The GP-SM model had the best predictive performance of 1.25 for RMSE and 27.96\% for MAPE. The GP-SM optimized model had a fairly similar predictive performance of 1.29 for RMSE and 28.80\% for the MAPE. The estimated ARIMA model had the worst predictive performance of 3.20 for RMSE and 93.44\% for the MAPE. These findings suggest that hyperparameter tuning for small data produces models of comparable performance. If parameter tuning takes a considerable amount of time, then it may not be necessarily to use if it does not produce any significant model improvement.

\begin{table}[h]
\caption{Performance measures for the three different models on email data}
\label{tab:performance-measures-emails}
\begin{center}
\begin{tabular}{l@{\qquad}cc}
  \toprule
  \multirow{2}{*}{\raisebox{-\heavyrulewidth}{\bf MODEL}} & \multicolumn{2}{c}{\bf MEASURE} \\
  \cmidrule{2-3}
  & {\bf RMSE} & {\bf MAPE} \\
  \midrule
  GP-SM & {\bf 1.25} & {\bf 27.96\%}  \\
  GP-SM Optimized & 1.29 & 28.80\%  \\
  ARIMA & 3.20 & 93.44\%  \\
  \bottomrule
\end{tabular}
\end{center}
\end{table}

\begin{figure*}[t]
      \subfloat[Number of fatal crashes per 100 million VMT\label{subfig:data-fatalities}]{%
       \includegraphics[width=3.1in]{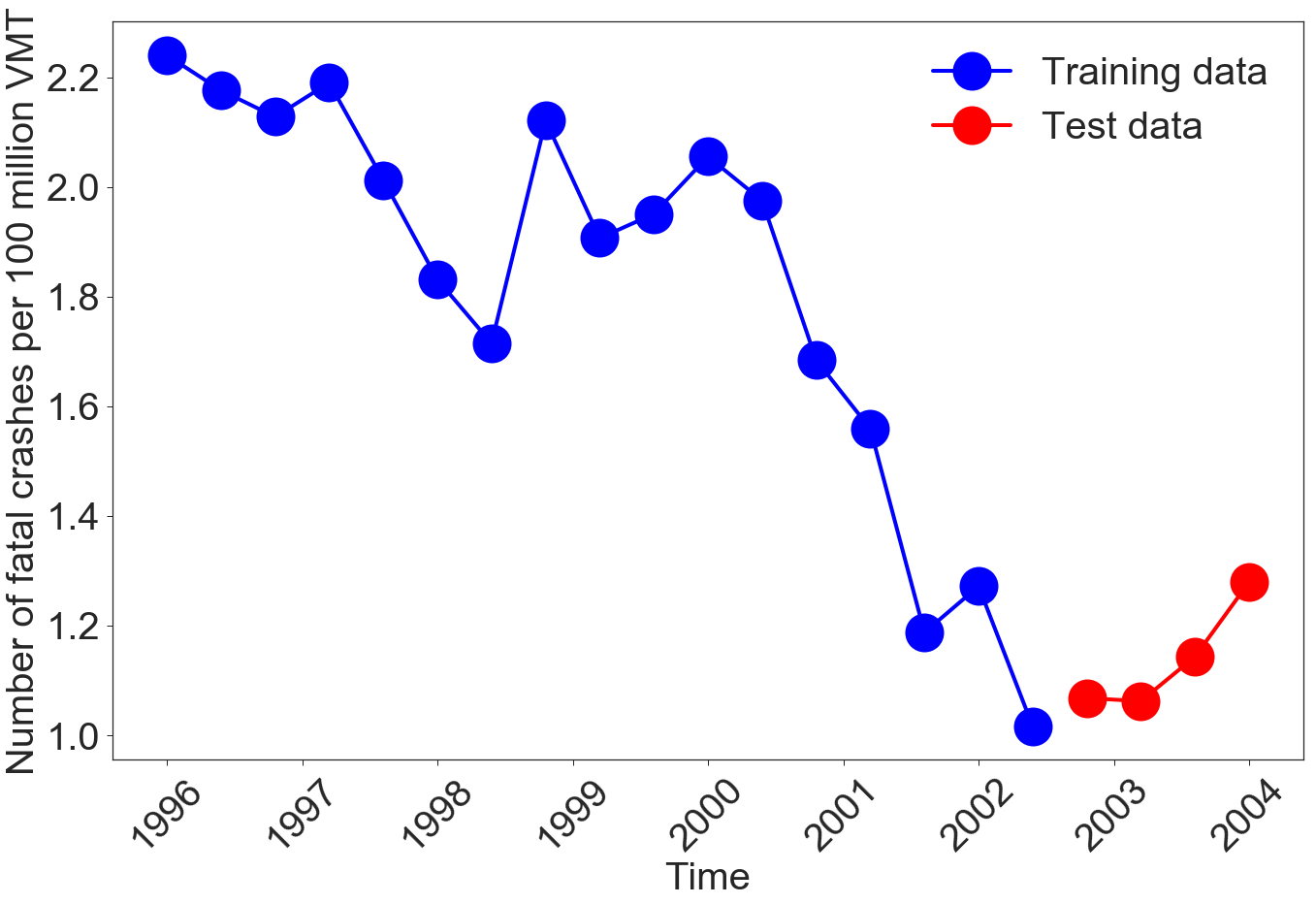}
     }  
     \hfill
     \subfloat[Number of serious injuries per 100 million VMT\label{subfig:data-injuries}]{%
       \includegraphics[width=3.1in]{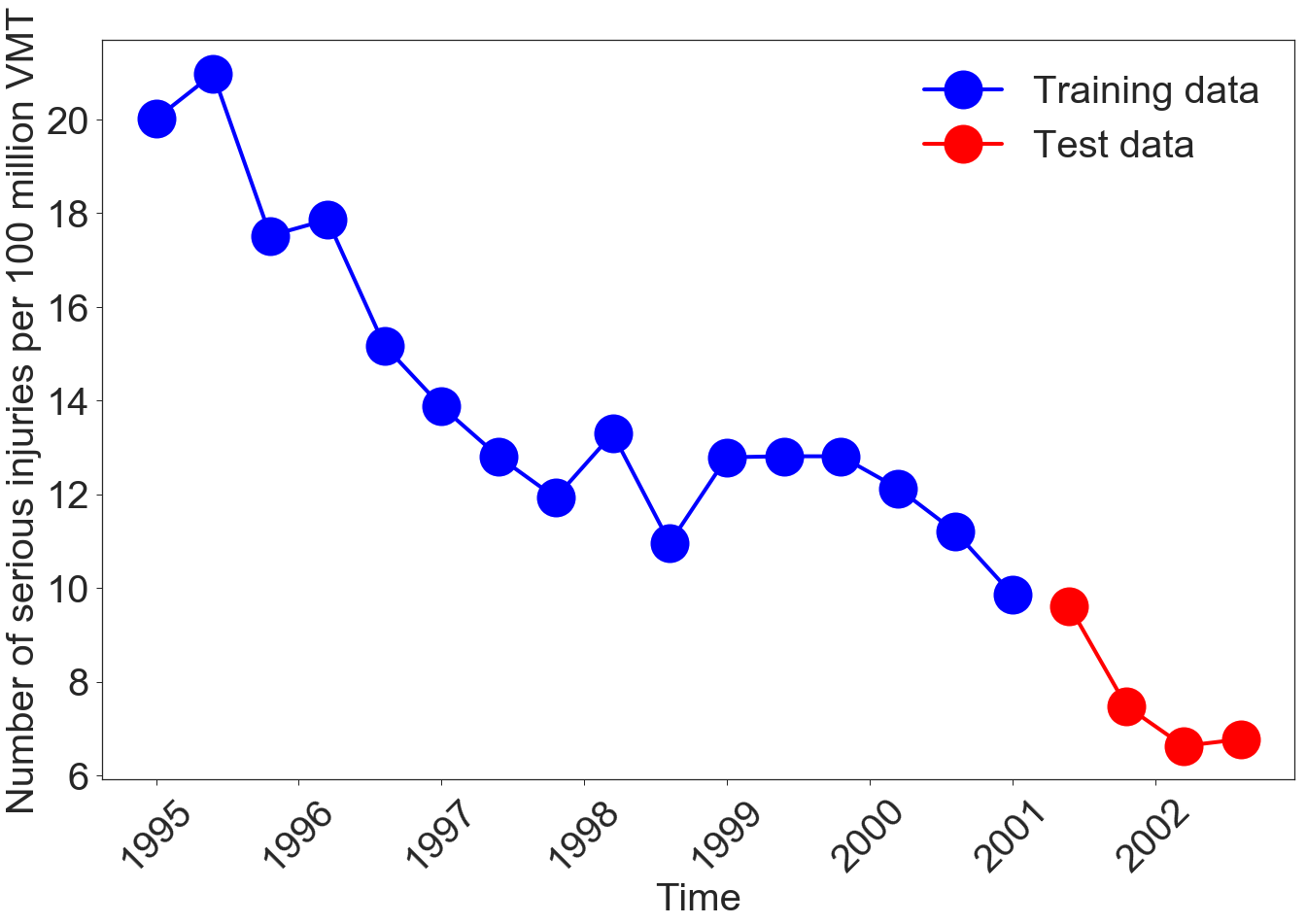}
     }   
     \caption{Number of fatal and serious injury crashes per 100 million vehicle miles travel.}
     \label{fig:data-crashes}
\end{figure*}

\begin{figure*}[t]
      \subfloat[GP-SM for fatalities\label{subfig:model-fatalities}]{%
       \includegraphics[width=0.48\linewidth]{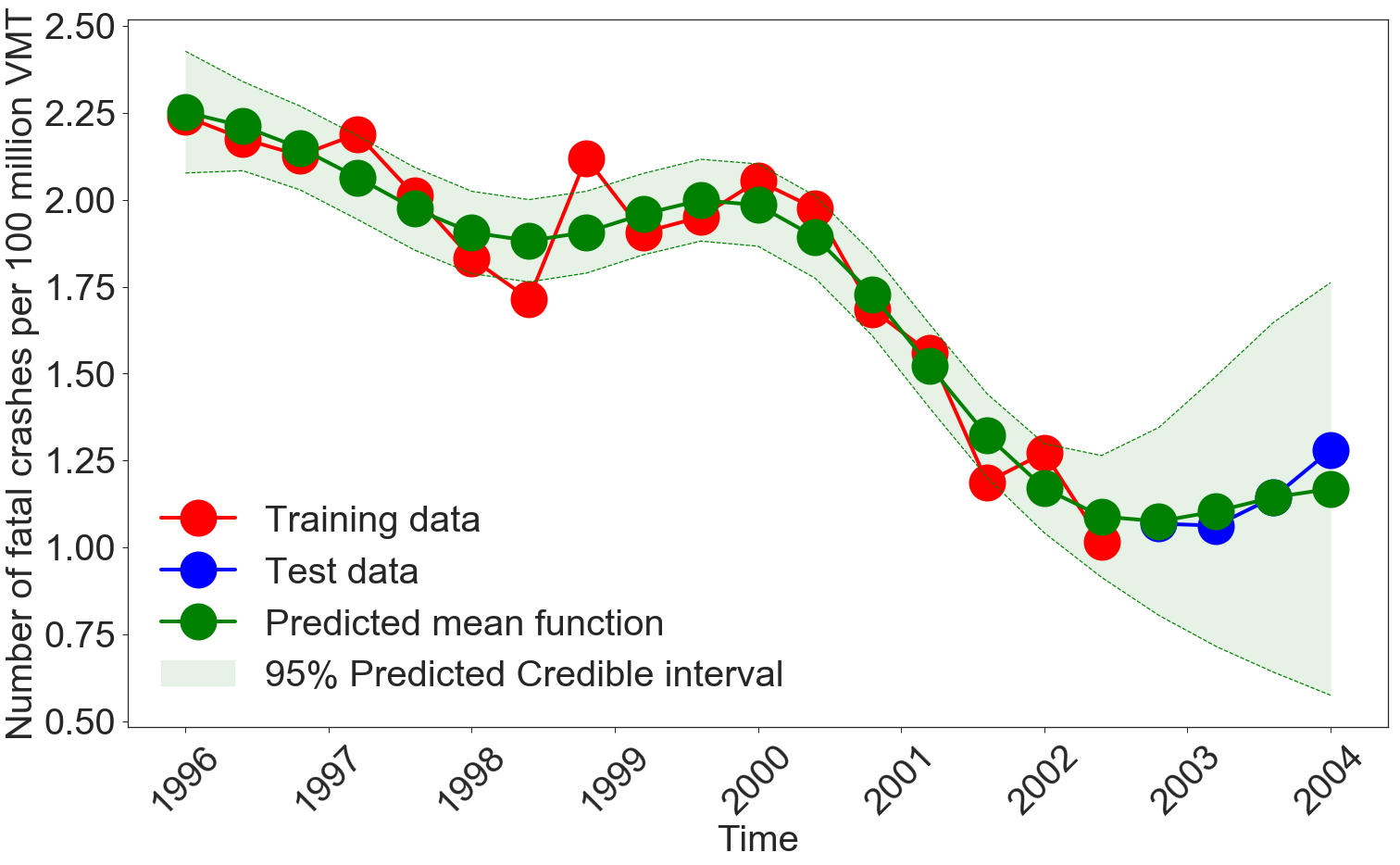}
     }
     \hfill
     \subfloat[GP-SM Optimized for fatalities\label{subfig:model-opt-fatalities}]{%
       \includegraphics[width=0.48\linewidth]{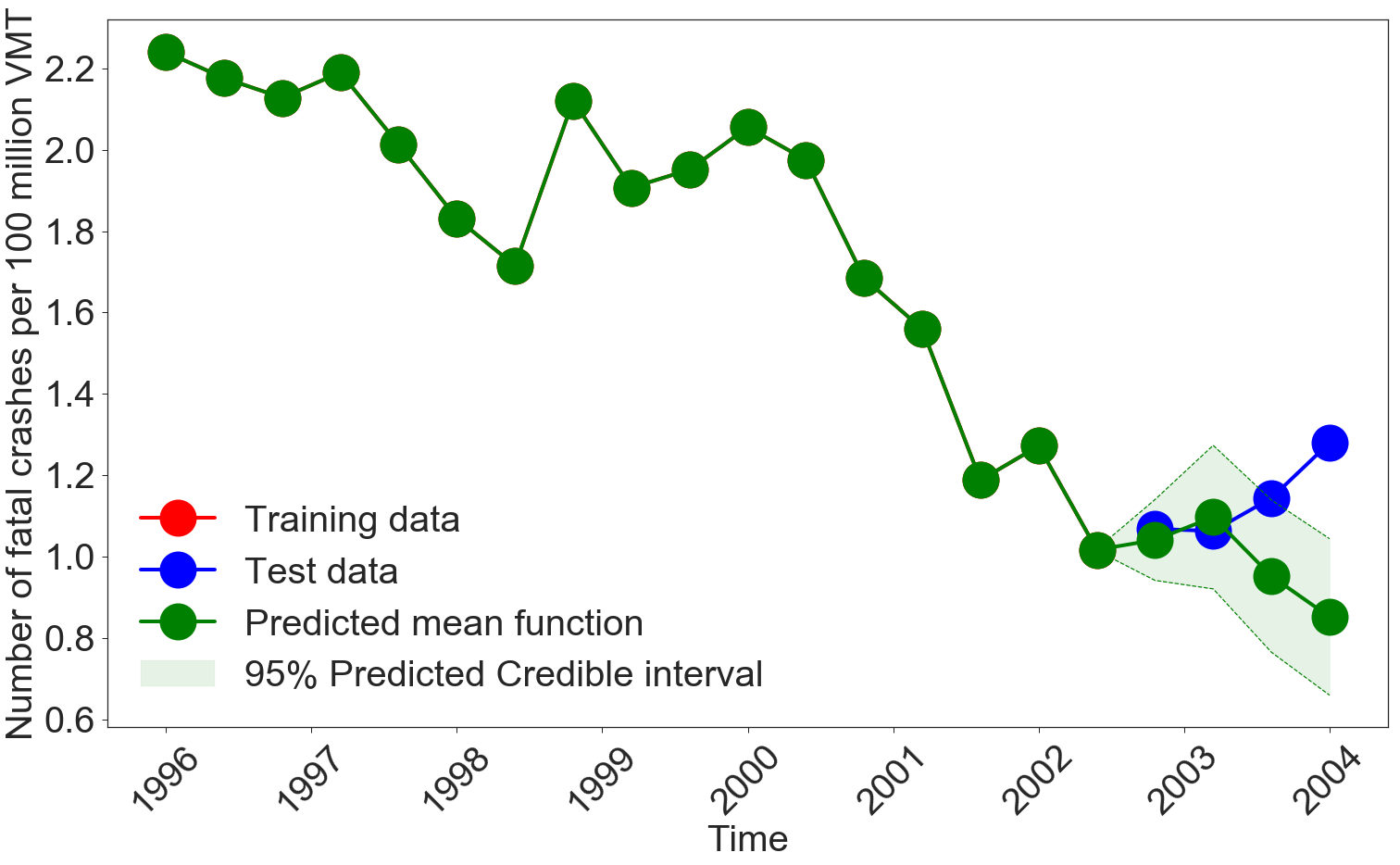}
     }
     \hfill
     \subfloat[ARIMA for fatalities\label{subfig:model-fatalities-arima}]{%
       \includegraphics[width=0.48\linewidth]{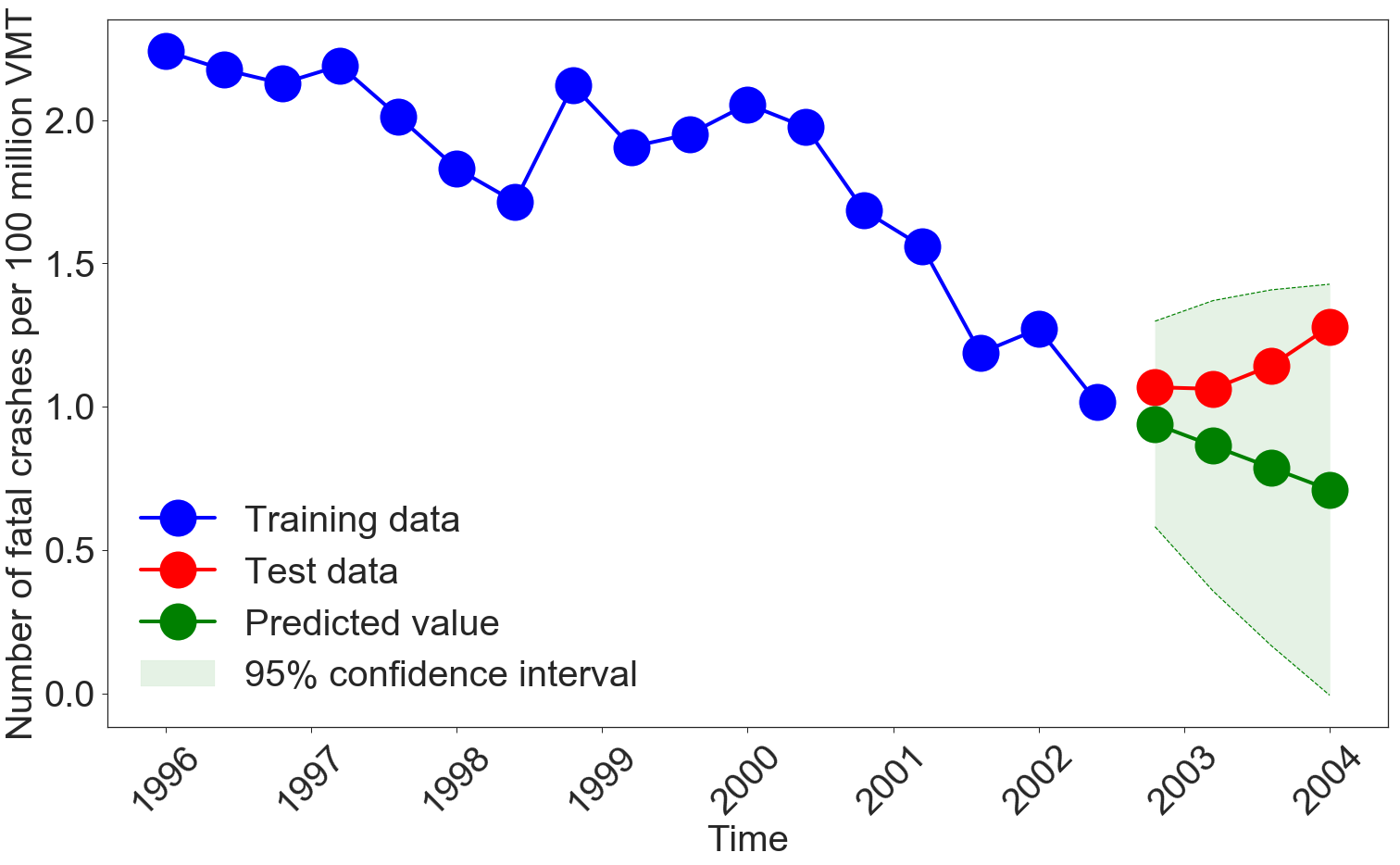}
     }
     \hfill
     \subfloat[Number of Iterations in Optimization for fatalities\label{subfig:iterations-fatalities}]{%
       \includegraphics[width=0.48\linewidth]{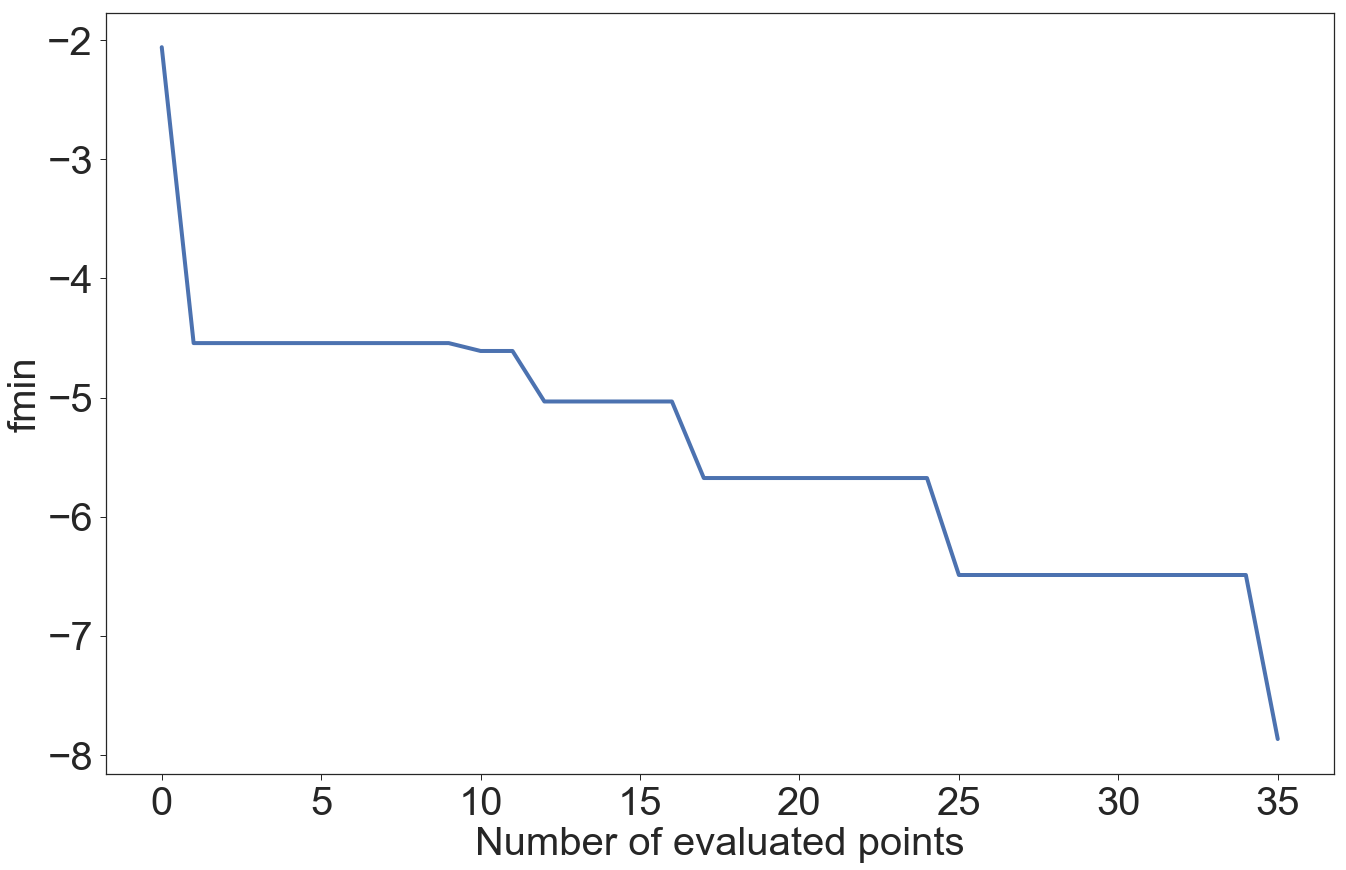}
     }
     \caption{Comparison of Gaussian process models with spectral mixture kernels to ARIMA for fatal crashes.}
     \label{fig:gp-posterior-fatalities}
\end{figure*}

\begin{figure*}[t]
      \subfloat[GP-SM for serious injuries\label{subfig:model-injuries}]{%
       \includegraphics[width=0.48\linewidth]{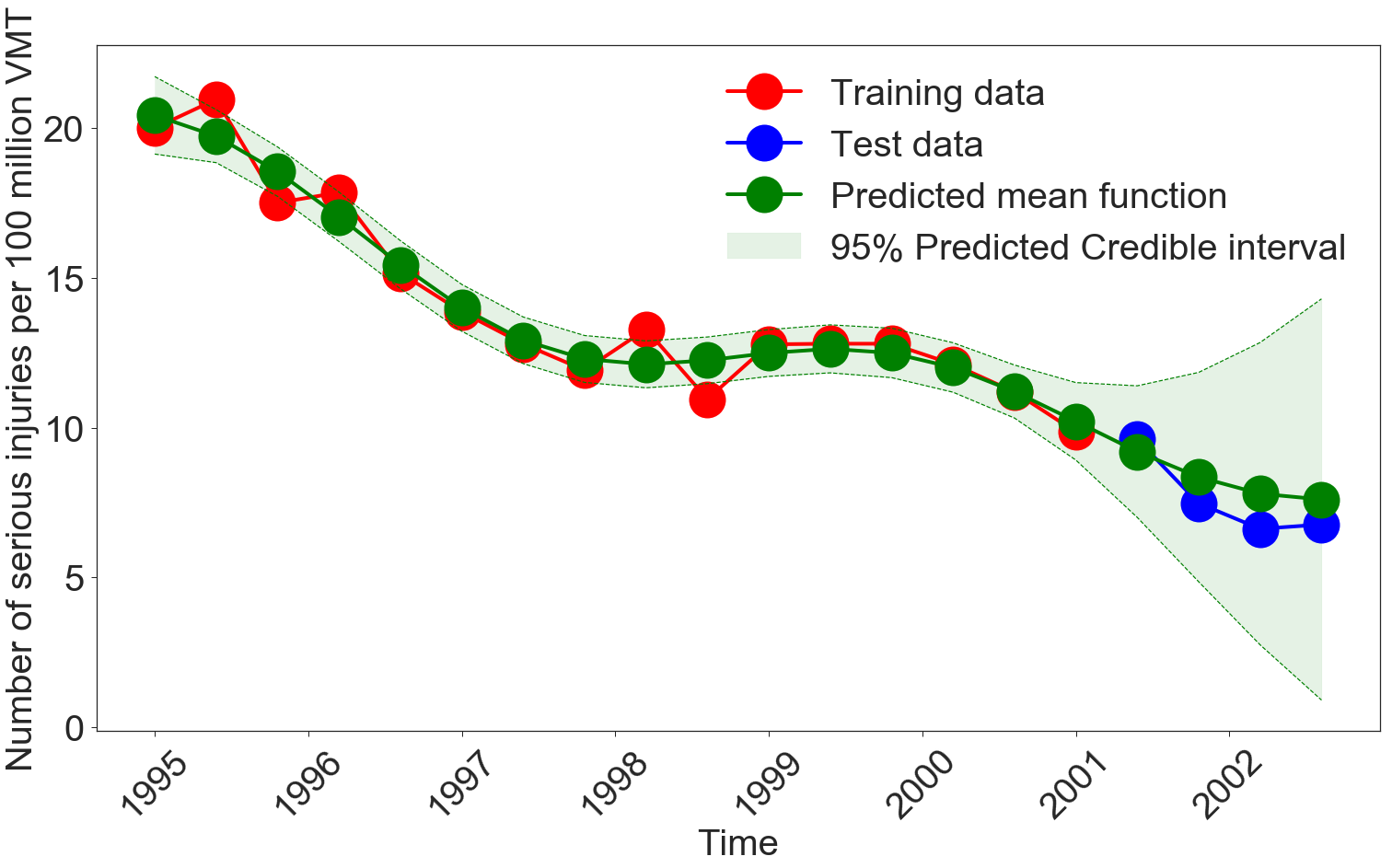}
     }
     \hfill
     \subfloat[GP-SM Optimized for serious injuries\label{subfig:model-opt-injuries}]{%
       \includegraphics[width=0.48\linewidth]{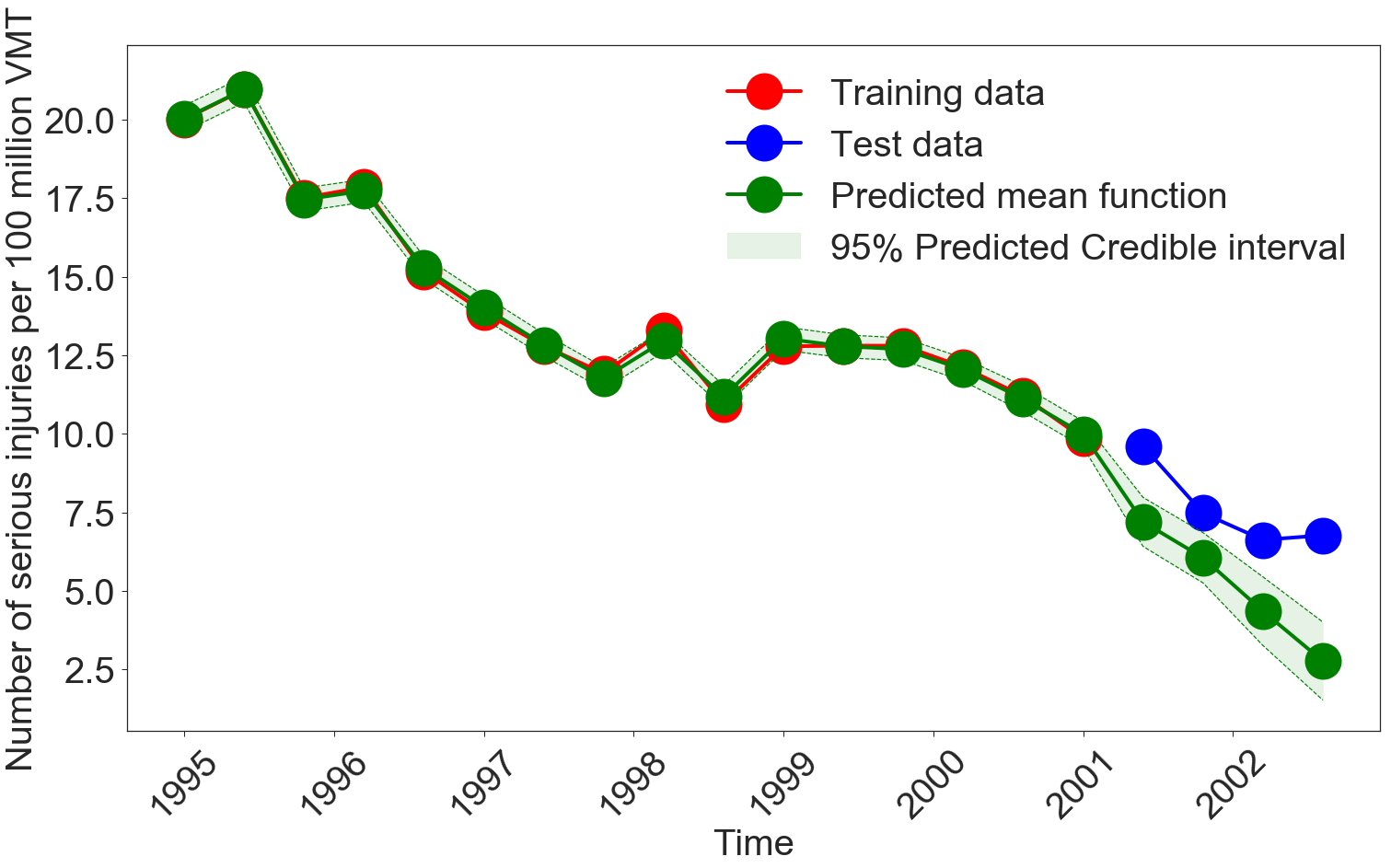}
     }
     \hfill
     \subfloat[ARIMA for serious injuries\label{subfig:model-injuries-arima}]{%
       \includegraphics[width=0.48\linewidth]{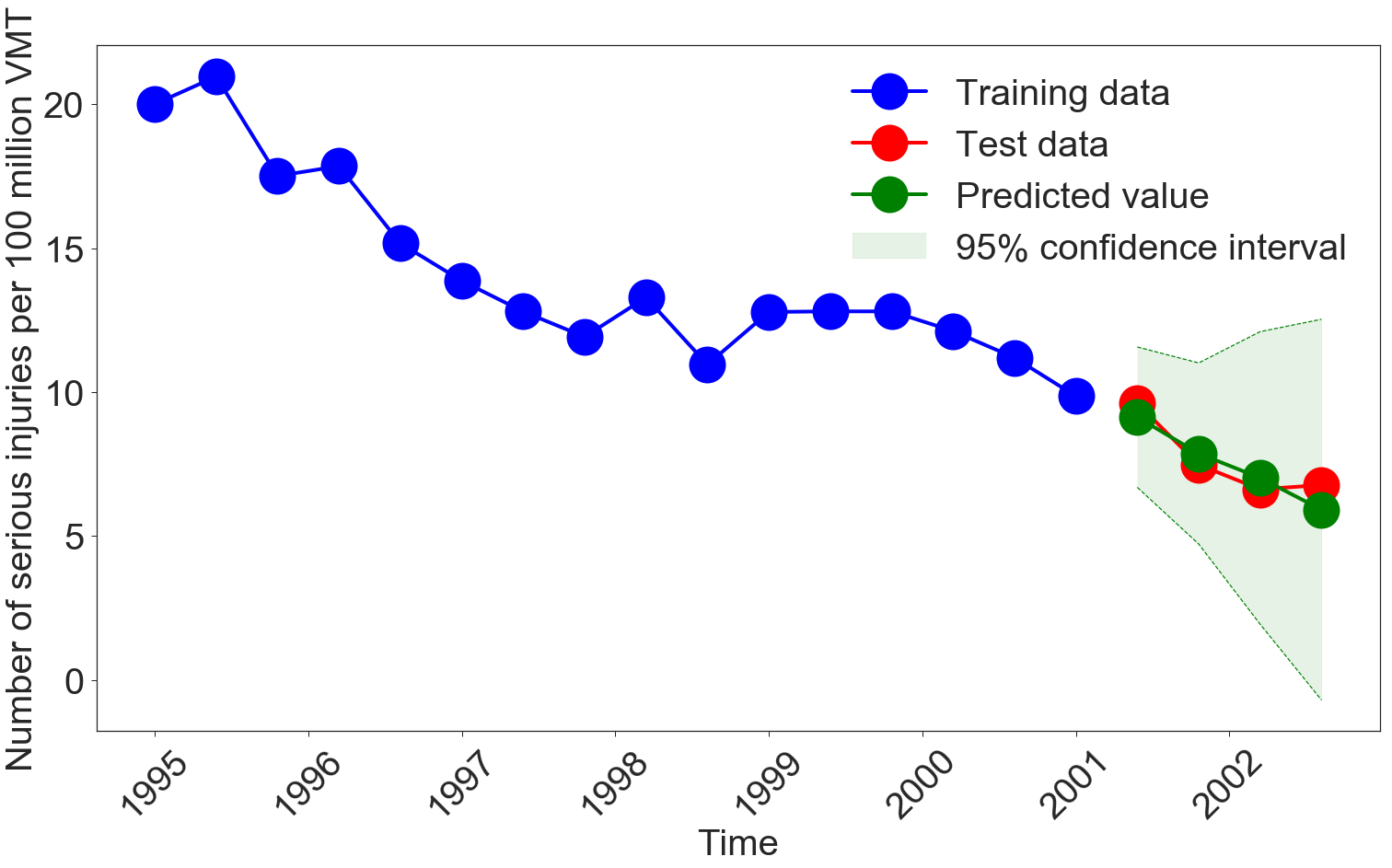}
     }
     \hfill
     \subfloat[Number of Iterations in Optimization for serious injuries\label{subfig:iterations-injuries}]{%
       \includegraphics[width=0.48\linewidth]{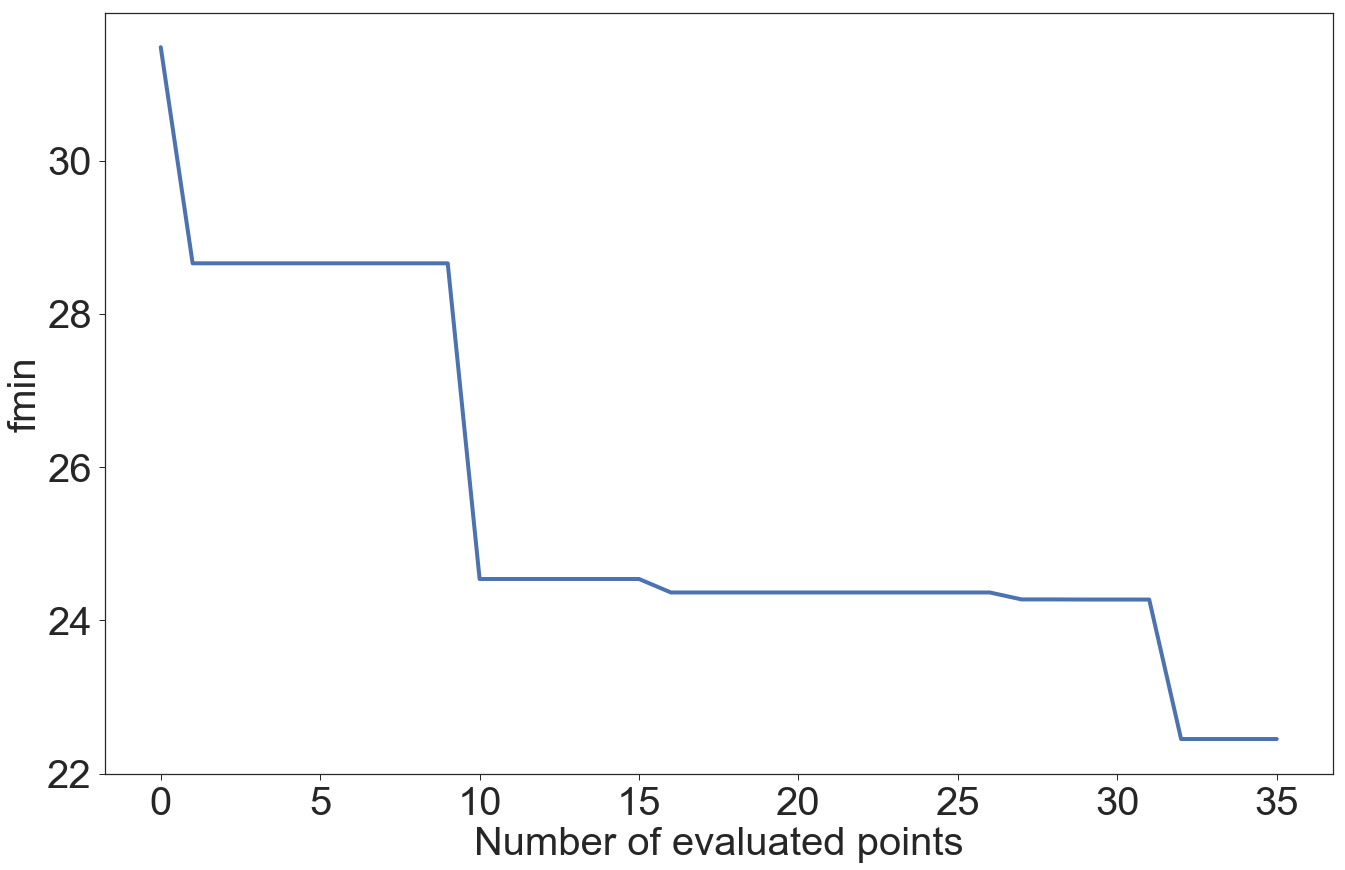}
     }    
     \caption{Comparison of Gaussian process models with spectral mixture kernels to ARIMA for serious injuries.}
     \label{fig:gp-posterior-injuries}
\end{figure*}

\subsection{HIGHWAY SAFETY}

\subsubsection{Raw data and sample formation}

The highway-crash data used for empirical analysis in this study was provided by the Nevada Department of Transportation (NDOT). This data consists of two types of crash severity including the annual number of fatalities and serious injuries standardized over an exposure variable called vehicle miles traveled (VMT). Vehicle miles traveled represents the total annual miles of vehicle travel divided by the total population in a state. Data for this indicator is provided by the Highway Statistics division of the Federal Highway Administration (FHWA) \citep{FHWA}. Data for fatal crashes and serious injuries were available from 1996 to 2004 and from 1995 to 2003, respectively. 

The final samples used for the current analysis include time-series of annual fatalities per 100 million VMT and annual serious injuries per 100 million VMT resulting in small sample sizes of 20 and 21 respectively. Table \ref{tab:descriptive-stats-fatalities} provides descriptive statistics of these time-series. Given that NDOT only collects these crash counts annually \citep{veeramisti2016business}, they lack seasonality as shown in Figure \ref{fig:data-crashes}. The first 17 and 16 data points shown in blue in Figure \ref{subfig:data-fatalities} and Figure \ref{subfig:data-injuries} were used for training, respectively. The remaining four data points shown in red were used for testing. 

\begin{table}[h]
\caption{Descriptive statistics for highway crashes per 100 million VMT in Nevada}
\label{tab:descriptive-stats-fatalities}
\begin{center}
\begin{tabular}{l@{\qquad}cc}
  \toprule
  \multirow{2}{*}{\raisebox{-\heavyrulewidth}{\bf STATISTIC}} & \multicolumn{2}{c}{\bf CRASH TYPE} \\
  \cmidrule{2-3}
  & {\bf Fatalities} & {\bf Serious~injuries} \\
  \midrule
  Mean & 1.69 & 12.82  \\
  Std Dev & 0.43 & 3.99  \\
  Min & 1.02 & 6.63  \\
  Max & 2.24 & 20.97  \\
  Count & 21 & 20  \\
  \bottomrule
\end{tabular}
\end{center}
\end{table}

% Citations within the text should include the author's last name and
% year, e.g., (Cheesman, 1985). Reference style should follow the style
% that you are used to using, as long as the citation style is
% consistent.
% For the original submission, take care not to reveal the authors' identity through
% the manner in which one's own previous work is cited.  For example, writing
% ``In (Bovik, 1970), we studied the problem of AI'' would be inappropriate, as
% it reveals the author's identity.  Instead, write ``(Bovik, 1970) studied the
% problem of AI.''

\subsubsection{Empirical Analysis}

An appropriate number of terms in the sum, $Q$, was set to 10 as proposed by \cite{wilson2015human} resulting in 10 frequency parameters and 10 length-scale parameters. A maximum number of 30 iterations was chosen to tune the 20 hyperparameters for the optimized GP-SM. An ARIMA model was estimated using the methodology proposed by \cite{veeramisti2016business} for comparison.

Figure \ref{fig:gp-posterior-fatalities} shows the results of the three estimated models for annual fatal crashes per 100M VMT. Figure \ref{subfig:model-fatalities} shows that the Gaussian process model with a spectral mixture kernel is able to capture the structure implicitly both in regions of the training and testing data. The 95\% predicted credible interval (CI) contains the true number of annual fatalities per 100M VMT albeit that they are are much wider than the other two models. Figure \ref{subfig:model-opt-fatalities} shows that the Gaussian process model with the hyperparameters of the spectral mixture kernel tuned using Bayesian optimization is also able to capture structure implicitly. The predictive performance is especially better in the region of the training data where the predicted data points entirely overlap with the training data. Predictive performance on testing data is poorer than the standard model. This finding suggests that hyperparameter tuning for small data without easily noticeable structure does not produce any significant model improvement. Figure \ref{subfig:model-fatalities-arima} shows that ARIMA model is not able to capture the structure within the region of testing data. This finding suggests that ARIMA models have poor performance for small data without noticeable structure. The 95\% confidence interval for ARIMA is much wider than both the GP models showing a high degree of uncertaininty about the ARIMA predictions.

Figure \ref{fig:gp-posterior-injuries} shows the results of the three estimated models for annual serious injuries per 100M VMT. Figure \ref{subfig:model-injuries} shows that the Gaussian process model with a spectral mixture kernel is able to capture the structure implicitly both in regions of the training and testing data. The 95\% predicted credible interval (CI) contains the true number of annual serious injuries per 100M VMT. The C.I are much wider than the other two models. Figure \ref{subfig:model-opt-injuries} shows that the Gaussian process model with the hyperparameters of the spectral mixture kernel tuned using Bayesian optimization is also able to capture structure implicitly. The predictive performance is especially better in the region of the training data where the predicted data points entirely overlap with the training data. Predictive performance in the region of testing data is poorer compared to the standard model. This finding suggests that hyperparameter tuning for small data without easily noticeable structure does not produce any significant model improvement. Figure \ref{subfig:model-injuries-arima} shows that ARIMA model was also able to capture the structure within the region of testing data. This finding suggests that ARIMA models have good performance for time-series that have easily noticeable structure. The 95\% confidence interval for ARIMA is comparable narrower than the standard GP model showing a lesser degree of uncertainty about the ARIMA predictions.

Table \ref{tab:performance-measures-emails} shows the performance measures for the three estimated models including the root mean square error (RMSE) and mean absolute percentage error (MAPE). The GP-SM model had the best predictive performance of 1.25 for RMSE and 27.96\% for MAPE. The GP-SM optimized model had a fairly similar predictive performance of 1.29 for RMSE and 28.80\% for the MAPE. The estimated ARIMA model had the worst predictive performance of 3.20 for RMSE and 93.44\% for the MAPE. These findings suggest that hyperparameter tunning for small data produces models of comparable performance. If parameter tuning takes a considerable amount of time, then it may not be necessarily to use if it does not produce any significant model improvement.

\begin{table}[h]
\caption{Performance measures for the three different models on highway crashes per 100 million VMT}
\label{tab:performance-measures-crashes}
\begin{center}
\begin{tabular}{l@{\quad}cc@{\quad}cc}
  \toprule
  \multirow{2}{*}{\raisebox{-\heavyrulewidth}{\bf MODEL}} & \multicolumn{2}{c}{\bf FATALITIES} & \multicolumn{2}{c}{\bf INJURIES} \\
  \cmidrule{2-5}
  & {\bf RMSE} & {\bf MAPE} & {\bf RMSE} & {\bf MAPE} \\
  \midrule
  GP-SM & {\bf 0.06} & {\bf 3.31\%} & {\bf 0.87} & {\bf 11.5\%}  \\
  GP-SM Opt & 0.24 & 14.00\% & 2.69 & 34.37\%  \\
  ARIMA & 0.34 & 27.07\% & 1.68 & 17.72\%  \\
  \bottomrule
\end{tabular}
\end{center}
\end{table}

\section{CONCLUSIONS AND FUTURE WORK}\label{conclusions-and-future-work}
This paper proposed a Bayesian nonparametric framework to capture implicitly hidden structure in time-series having limited data. The proposed framework, a Gaussian process with a spectral mixture kernel, was applied to time-series process for insider-threat detection and highway-safety planning. The proposed framework addresses two current challenges when analyzing quite noisy time-series having limited data whereby the time series are visualized for noticeable structure such as periodicity, growing or decreasing trends and hard coding them into pre-specified functional forms. Experiments demonstrated that results from this framework outperform traditional ARIMA when the time series does not have easily noticeable structure and is quite noisy. Future work will involve evaluating the proposed framework on other different types of insider-threat behavior. 

% \subsubsection{Footnotes}

% Indicate footnotes with a number\footnote{Sample of the first
% footnote} in the text. Use 8 point type for footnotes.  Place the
% footnotes at the bottom of the page on which they appear.  Precede the
% footnote with a 0.5 point horizontal rule 1~inch (6~picas)
% long.\footnote{Sample of the second footnote}

% \subsubsection{Figures}

% All artwork must be centered, neat, clean, and legible. Figure number
% and caption always appear below the figure.  Leave 2 line spaces
% between the figure and the caption. The figure caption is initial caps
% and each figure numbered consecutively.

% Make sure that the figure caption does not get separated from the
% figure. Leave extra white space at the bottom of the page rather than
% splitting the figure and figure caption.

% \subsubsection{Tables}

% All tables must be centered, neat, clean, and legible. Table number
% and title always appear above the table.  See
% Table~\ref{sample-table}.

% One line space before the table title, one line space after the table
% title, and one line space after the table. The table title must be
% initial caps and each table numbered consecutively.

% \begin{table}[h]
% \caption{Sample Table Title}
% \label{sample-table}
% \begin{center}
% \begin{tabular}{ll}
% \multicolumn{1}{c}{\bf PART}  &\multicolumn{1}{c}{\bf DESCRIPTION} \\
% \hline \\
% Dendrite         &Input terminal \\
% Axon             &Output terminal \\
% Soma             &Cell body (contains cell nucleus) \\
% \end{tabular}
% \end{center}
% \end{table}

% \newpage

\subsubsection*{Acknowledgements}

The authors would like to thank the CERT-Division and Dr. Naveen Veeramisti for providing the "Insider Threat Tools dataset" and NDOT's highway-crash data, respectively, used in the empirical analyses. 
% Special thanks to David Jones, Bob Schrag, and Rob Kerr for providing critical comments. Special thanks to John Boatman for proof reviewing earlier versions of this paper.

% \subsubsection*{References}

% References follow the acknowledgements.  Use unnumbered third level
% heading for the references title.  Any choice of citation style is
% acceptable as long as you are consistent.

\bibliographystyle{apa}
\bibliography{mybibfile}

\begin{thebibliography}{}

\bibitem[\protect\astroncite{Brochu et~al.}{2010}]{brochu2010tutorial}
Brochu, E., Cora, V.~M., and De~Freitas, N. (2010).
\newblock A tutorial on bayesian optimization of expensive cost functions, with
  application to active user modeling and hierarchical reinforcement learning.
\newblock {\em arXiv preprint arXiv:1012.2599}.

\bibitem[\protect\astroncite{CERT-Division}{}]{CERT}
CERT-Division.
\newblock {Software Engineering Institute}.
\newblock \url{https://www.sei.cmu.edu/about/divisions/cert/index.cfm}.
\newblock Accessed: January 30, 2018.

\bibitem[\protect\astroncite{Duvenaud}{2014}]{duvenaud2014automatic}
Duvenaud, D. (2014).
\newblock {\em Automatic model construction with Gaussian processes}.
\newblock PhD thesis, University of Cambridge.

\bibitem[\protect\astroncite{Emaasit and Paz}{2018}]{emaasit2018simultaneous}
Emaasit, D. and Paz, A. (2018).
\newblock Simultaneous estimation of flexible models and associated
  hyperparameters: An application to activity-duration modeling.
\newblock In {\em Transportation Research Board 97th Annual Meeting. Washington
  DC: Transportation Research Board}.

\bibitem[\protect\astroncite{FHWA}{}]{FHWA}
FHWA.
\newblock {Highway Statistics Series}.
\newblock \url{https://www.fhwa.dot.gov/policyinformation/statistics.cfm}.
\newblock Accessed: January 30, 2018.

\bibitem[\protect\astroncite{Ghahramani}{2015}]{ghahramani2015probabilistic}
Ghahramani, Z. (2015).
\newblock Probabilistic machine learning and artificial intelligence.
\newblock {\em Nature}, 521(7553):452--459.

\bibitem[\protect\astroncite{Gheyas and Abdallah}{2016}]{gheyas2016detection}
Gheyas, I.~A. and Abdallah, A.~E. (2016).
\newblock Detection and prediction of insider threats to cyber security: a
  systematic literature review and meta-analysis.
\newblock {\em Big Data Analytics}, 1(1):6.

\bibitem[\protect\astroncite{Glasser and Lindauer}{2013}]{glasser2013bridging}
Glasser, J. and Lindauer, B. (2013).
\newblock Bridging the gap: A pragmatic approach to generating insider threat
  data.
\newblock In {\em Security and Privacy Workshops (SPW), 2013 IEEE}, pages
  98--104. IEEE.

\bibitem[\protect\astroncite{Greitzer and Ferryman}{2013}]{greitzer2013methods}
Greitzer, F.~L. and Ferryman, T.~A. (2013).
\newblock Methods and metrics for evaluating analytic insider threat tools.
\newblock In {\em Security and Privacy Workshops (SPW), 2013 IEEE}, pages
  90--97. IEEE.

\bibitem[\protect\astroncite{Hjort et~al.}{2010}]{hjort2010bayesian}
Hjort, N.~L., Holmes, C., M{\"u}ller, P., and Walker, S.~G. (2010).
\newblock {\em Bayesian nonparametrics}, volume~28.
\newblock Cambridge University Press.

\bibitem[\protect\astroncite{Hoffman and Gelman}{2014}]{hoffman2014no}
Hoffman, M.~D. and Gelman, A. (2014).
\newblock The no-u-turn sampler: adaptively setting path lengths in hamiltonian
  monte carlo.
\newblock {\em Journal of Machine Learning Research}, 15(1):1593--1623.

\bibitem[\protect\astroncite{Kucukelbir et~al.}{2015}]{kucukelbir2015automatic}
Kucukelbir, A., Ranganath, R., Gelman, A., and Blei, D. (2015).
\newblock Automatic variational inference in stan.
\newblock In {\em Advances in neural information processing systems}, pages
  568--576.

\bibitem[\protect\astroncite{Kweon and Lim}{2012}]{kweon2012appropriate}
Kweon, Y.-J. and Lim, I.-K. (2012).
\newblock Appropriate regression model types for intersections in
  safetyanalyst.
\newblock {\em Journal of Transportation Engineering}, 138(10):1250--1258.

\bibitem[\protect\astroncite{Lindauer et~al.}{2014}]{lindauer2014generating}
Lindauer, B., Glasser, J., Rosen, M., Wallnau, K.~C., and ExactData, L. (2014).
\newblock Generating test data for insider threat detectors.
\newblock {\em JoWUA}, 5(2):80--94.

\bibitem[\protect\astroncite{McHutchon et~al.}{2014}]{mchutchon2014nonlinear}
McHutchon, A. et~al. (2014).
\newblock {\em Nonlinear modelling and control using Gaussian processes}.
\newblock PhD thesis, Citeseer.

\bibitem[\protect\astroncite{Schulz et~al.}{2017}]{schulz2017compositional}
Schulz, E., Tenenbaum, J.~B., Duvenaud, D., Speekenbrink, M., and Gershman,
  S.~J. (2017).
\newblock Compositional inductive biases in function learning.
\newblock {\em Cognitive psychology}, 99:44--79.

\bibitem[\protect\astroncite{Smith}{2016}]{smith2016hsip}
Smith, S. (2016).
\newblock {\em HSIP 2016 National Summary Report}.
\newblock Number FHWA-SA-17-040.

\bibitem[\protect\astroncite{Stoffel et~al.}{2013}]{stoffel2013finding}
Stoffel, F., Fischer, F., and Keim, D.~A. (2013).
\newblock Finding anomalies in time-series using visual correlation for
  interactive root cause analysis.
\newblock In {\em Proceedings of the Tenth Workshop on Visualization for Cyber
  Security}, pages 65--72. ACM.

\bibitem[\protect\astroncite{Sukhai et~al.}{2011}]{sukhai2011temporal}
Sukhai, A., Jones, A.~P., Love, B.~S., and Haynes, R. (2011).
\newblock Temporal variations in road traffic fatalities in south africa.
\newblock {\em Accident Analysis \& Prevention}, 43(1):421--428.

\bibitem[\protect\astroncite{Veeramisti}{2016}]{veeramisti2016business}
Veeramisti, N.~K. (2016).
\newblock {\em A business intelligence framework for network-level traffic
  safety analyses}.
\newblock PhD thesis, University of Nevada, Las Vegas.

\bibitem[\protect\astroncite{Williams and
  Rasmussen}{2006}]{williams2006gaussian}
Williams, C.~K. and Rasmussen, C.~E. (2006).
\newblock Gaussian processes for machine learning.
\newblock {\em the MIT Press}, 2(3):4.

\bibitem[\protect\astroncite{Wilson and Adams}{2013}]{wilson2013gaussian}
Wilson, A. and Adams, R. (2013).
\newblock Gaussian process kernels for pattern discovery and extrapolation.
\newblock In {\em International Conference on Machine Learning}, pages
  1067--1075.

\bibitem[\protect\astroncite{Wilson}{2014}]{wilson2014covariance}
Wilson, A.~G. (2014).
\newblock Covariance kernels for fast automatic pattern discovery and
  extrapolation with gaussian processes.
\newblock {\em University of Cambridge}.

\bibitem[\protect\astroncite{Wilson et~al.}{2015}]{wilson2015human}
Wilson, A.~G., Dann, C., Lucas, C., and Xing, E.~P. (2015).
\newblock The human kernel.
\newblock In {\em Advances in neural information processing systems}, pages
  2854--2862.

\bibitem[\protect\astroncite{Yannis et~al.}{2011}]{yannis2011modeling}
Yannis, G., Antoniou, C., and Papadimitriou, E. (2011).
\newblock Modeling traffic fatalities in europe.
\newblock In {\em 90th Annual Meeting of the Transportation Research Board,
  Washington, DC}.

\end{thebibliography}

% J.~Alspector, B.~Gupta, and R.~B.~Allen  (1989). Performance of a
% stochastic learning microchip.  In D. S. Touretzky (ed.), {\it Advances
% in Neural Information Processing Systems 1}, 748-760.  San Mateo, Calif.:
% Morgan Kaufmann.

% F.~Rosenblatt (1962). {\it Principles of Neurodynamics.} Washington,
% D.C.: Spartan Books.

% G.~Tesauro (1989). Neurogammon wins computer Olympiad.  {\it Neural
% Computation} {\bf 1}(3):321-323.

\end{document}